%% file: acl_latex.tex
\documentclass[11pt]{article}

\usepackage[]{emnlp2022}

\usepackage{times}
\usepackage{latexsym}

\usepackage[T1]{fontenc}
\usepackage[utf8]{inputenc}
\DeclareUnicodeCharacter{0307}{u}
\usepackage{microtype}

\usepackage{enumitem}
\usepackage{graphics}
\usepackage{todonotes} % Insert [disable] to disable all notes:

\input{mathabbrev}

\title{He Said, She Said: Style Transfer for\\ Shifting the Perspective of Dialogues}

\author{Amanda Bertsch \\
  Carnegie Mellon University \\
  \texttt{abertsch@cs.cmu.edu} \\\And
  Graham Neubig \\
  Carnegie Mellon University \\
  \texttt{gneubig@cs.cmu.edu}\\\And
  Matthew R. Gormley \\
  Carnegie Mellon University \\
  \texttt{mgormley@cs.cmu.edu} \\}

\begin{document}
\maketitle
\begin{abstract}
In this work, we define a new style transfer task: perspective shift, which reframes a dialouge from informal first person to a formal third person rephrasing of the text. 
This task requires challenging coreference resolution, emotion attribution, and interpretation of informal text. 
We explore several baseline approaches and discuss further directions on this task when applied to short dialogues. 
As a sample application, we demonstrate that applying perspective shifting to a dialogue summarization dataset (SAMSum) substantially improves the zero-shot performance of extractive news summarization models on this data. 
Additionally, supervised extractive models perform better when trained on perspective shifted data than on the original dialogues.
We release our code publicly.\footnote{\url{https://github.com/abertsch72/perspective-shifting}}
\end{abstract}

\section{Introduction}

Style transfer models change surface attributes of text while preserving the content. 
Previous work on style transfer has focused on controlling the formality, authorial style, and sentiment of text \cite{jin-etal-2022-deep}. 
We propose a new style transfer task: perspective shift from dialogue to 3rd person conversational accounts (\S \ref{sec:task-definition}).
In this task, we seek to convert from an informal 1st person transcription of the dialogue to a 3rd person rephrasing of the conversation, where each line captures the information of a single utterance with relevant contextualizing information added.
Table \ref{tab:exampleconvo} demonstrates an example conversion and its perspective shifted version.

This task is challenging because it requires the interpretation of many discourse phenomena.
In dialogue, speakers commonly use 1st and 2nd person pronouns and casual speech. 
Speakers also convey their own emotions and opinions in their speech. 
Converting a multi-party conversation to a single-perspective rephrasing requires pronoun resolution, formalization, and attribution of emotion/stance markers to individuals. 
While coreference resolution, stance detection, and formalization are often treated as separate tasks, the signal for these objectives is commingled in the dialogues. 
A pipeline approach would discard information necessary for any one task in the completion of the other two.

We create a dataset for this task by annotating dialogues from the SAMSum corpus \cite{gliwa-etal-2019-samsum}, a dialogue summarization corpus of synthetic text message conversations (\S \ref{sec:dataset-creation}). 
For each conversation, annotators rephrase the utterances line-by-line into one or more sentences in 3rd person. 
Unlike a summary, which condenses information to highlight the most important points, the goal of this transformation is to capture as much of the information from the original utterance as possible in a more standardized form.

We fine-tune BART on this dataset as a supervised baseline under several different problem formulations, and we experiment with incorporating formality data into the training process (\S \ref{sec:perspective-shifting}).
As a motivating use case, we demonstrate that extractive summarization over perspective-shifted dialogue is more fluent and has higher ROUGE scores than extractive summarization over the original dialogues (\S \ref{sec:extractive}). 
This trend holds for zero-shot performance of extractive summarization models trained on news corpora and for fully supervised training on model-generated perspective shift data. 

Perspective shift can be a useful operation for extractive summarization when annotation time is limited; when additional data from out-of-domain is available; when the exact length and content of the summary is not known at annotation time; or when high faithfulness is important to the end task, but fluency is also a concern (\S  \ref{sec:benefits-of-ps}).

\begin{table*}[t]
\begin{tabular}{p{0.42\linewidth} | p{0.56\linewidth}}

Original                                    & Perspective shifted                                        \\ \hline
Laura: I need a new printer :/              & Laura is frustrated that she needs a new printer.          \\
Laura: thinking about this one              & Laura is thinking about a specific printer.                \\
Laura: \textless{}file\_other\textgreater{} & Laura sends a file.                                        \\
Jamie: you're sure you need a new one?      & Jamie asks if Laura is sure she needs a new one.           \\
Jamie: I mean you can buy a second hand one & Jamie clarifies that Laura could buy a secondhand printer. \\
Laura: could be                             & Laura says that's possible.                               
\end{tabular}
\caption{An example conversation from the SAMSum dataset with the associated perspective shift.}
\label{tab:exampleconvo}

\end{table*}

\section{Task definition}
\label{sec:task-definition}
We define \emph{perspective shift} as an utterance-level rephrasing task. 
Given a dialogue and a single selected utterance, the goal of the task is to rewrite that utterance as a formal third person statement. 
Four operations are required to accomplish this change: coreference resolution, syntactic rewriting, formalization, and emotion attribution. 
Table \ref{tab:exampleconvo} shows an example conversation and perspective shift, demonstrating each of these challenges. 

First-person singular and second-person pronouns are usually easily resolved in a conversational context---first-person singular refers to the speaker, while second-person pronouns generally refer to the other conversational parties---plural first-person pronouns can be less obvious to resolve. 
When a party in a conversation uses the pronoun ``we,'' this plural may be referring to the other parties in the conversation, some but not all of the parties in the conversation, or a party not present in the conversation, e.g. in the utterance ``I need to talk to my husband. We might have other plans.'' 
In our hand-annotated dataset, we resolve these pronouns wherever possible; if it is not clear what group the pronoun refers to, we resolve the pronoun as referring to ``<the current speaker> and others,'' e.g. ``Laura: we are busy'' becomes ``Laura and others are busy''. 
Other entities in the text may also be difficult to resolve, such as those defined only at the beginning of the conversation, many turns prior to the current reference.

Syntactic rewriting is the problem of converting the syntax of the utterance to reflect 3rd rather than 1st person.
This may involve re-conjugating verbs, e.g. converting ``Sam: I \textit{am} busy'' to ``Sam \textit{is} busy.'' 

% we conjecture
Formalization and emotion attribution are related problems, as much of the emotion and stance information in the text is contained in informal phrases, unconventional punctuation, and emojis \cite{heteroglossia}.
Typical formalization eliminates these markers without replacement \cite{rao-tetreault-2018-dear}.
However, this makes formalization a highly lossy conversion, which may be undesirable for downstream tasks.
We aim to limit the information lost in the perspective shift operation by encoding the meanings of such informal language in the output.
Often this takes the form of an adverb (e.g. ``Sam angrily says'') or a short descriptive sentence (e.g. ``Cam is amused''). 
This requires interpretation of the informal elements of the text.

Clearly, this task is far more complex than simply swapping pronouns for speaker names. 
We curate a dataset for the perspective shift operation.
\section{Dataset creation}
\label{sec:dataset-creation}

The dataset is an annotated subset of the SAMSum \cite{gliwa-etal-2019-samsum} dataset for dialogue summarization.
SAMSum is a dataset of simulated text message conversations, ranging from 3 to 30 lines in length and with between 2 and 20 speakers.
The dataset consists of 314 conversations from the train set, 368 conversations from the validation set, and 151 conversations from the test set \footnote{Due to SAMSum’s restrictive licensing, we are unable to release the dataset at this time. The SAMSum authors did not approve our requested exception.}. 
We set aside the 151 conversations from test as a test split and use the other 682 conversations as training and validation data. 

Annotators were instructed to convert each utterance individually to a formal 3rd person rephrasing, while preserving as much of the tone of the utterance as possible. 
Annotators were required to insert the speaker's name in each rewritten utterance and remove all 1st-person pronouns. 
Annotators were also asked to standardize grammar, remove questions, and add additional context (e.g. descriptive adverbs) to convey emotions previously expressed by emoticons. 
Further information about annotator selection and pay, as well as a full copy of the annotation instructions, is available in Appendix \ref{sec:appendix-annotation}.

\begin{table*}[t]
\centering
\begin{tabular}{l|llll}

Method             & ROUGE-1 & ROUGE-2 & ROUGE-L & BARTScore \\ \hline
\textit{no context}         &   62.57      &    40.45     &    61.41    &     -2.38    \\
\textit{left context only}  & 60.80        &  37.50       &  59.27       &  -2.39         \\
\textit{left and right context} &  \textbf{63.57}       &   \textbf{40.74}     &     \textbf{62.04 }   &   \textbf{-2.36 }      \\ 
\textit{conversation-level} & 63.20 & 35.04 & 51.80 &    -2.67       
\end{tabular}
\caption{Scores on the test set for models trained with different problem formulations.}
\label{tab:input-types}
\end{table*}

\subsection{Dataset statistics}
The perspective shifted conversations differ from the original in several ways. 
The number of turns in each conversation is preserved, but the average turn length varies: for the perspective shifts, the mean number of words per turn is 11.0, while the mean for the original dialogues is 8.4. 
(Note that the simplest heuristic would increase each utterance's word count by 1, as the colon next to the speaker name is swapped out with the word ``says''). \\

The average word-wise edit distance between original and perspective-shifted utterances is 8.5 words. 
This is partially due to the insertion of a dialogue tag (e.g. ``says'') in each utterance, the removal of emojis (average 0.1 per utterance), and the resolving of first and second person pronouns (average 0.9 per utterance). 
The part of speech\footnote{Part-of-speech related statistics are calculated using the spaCy POS tagger \cite{spacy}.} distribution of the conversations also changes, with a strong (65.8\%) decrease in interjections and a slight (5.1\%) decrease in adjectives and adverbs. 
However, in utterances that contain at least one emoji, the number of adjectives and adverbs present increases 12.8\%.
This is consistent with the annotation guidelines, which instruct annotators to capture the meaning of informal markers such as emoji with descriptors.

\section{Perspective shifting}
\label{sec:perspective-shifting}

\subsection{Formulation of the Prediction Problem}
\label{sec:input-data-manipulation}

\paragraph{Methods}

We consider several formulations of the perspective shifting task as a prediction problem with different input and output styles. 
Below, the first three approaches formulate the problem as a line-by-line task: each input example consists of the full conversation with one utterance designated as the utterance to be perspective shifted. The fourth approach below formulates the problem as conversation-level task in which the entire conversation is perspective shifted at once.
\begin{enumerate}
    \item \textit{no context:} The input to the model is the utterance $\uv_t$, and the output is the perspective shifted version, $\yv_t$.  
    \item \textit{left context only:} The input is the dialogue up to and including utterance $\uv_t$, and the output is the perspective shifted version, $\yv_t$. A \texttt{[SEP]} token delimits the left context, $\uv_1, \ldots, \uv_{t-1}$, from the utterance $\uv_t$. 
    \item \textit{left and right context:} The input is the full conversation, with \texttt{[SEP]} tokens around the utterance $\uv_t$, and the output is the perspective shifted version, $\yv_t$.
    \item \textit{conversation-level:} The input is a complete dialogue  $\uv_1, \ldots, \uv_T$, and the output is a complete perspective shift $\yv_1, \ldots, \yv_T$.
\end{enumerate}
For each formulation, we finetune a BART-large \cite{devlin-etal-2019-bert} model for 15 epochs, using early stopping, an effective batch size of 8, and a learning rate of 5e-5.

\paragraph{Results}

ROUGE 1/2/L scores and BARTScore for each model are listed in Table \ref{tab:input-types}.

The \emph{no context} model treats this as a purely utterance-level task, but fully precludes the addition of context from other utterances. 
This means that second-person and first-person plural pronouns cannot be resolved clearly. 
While this model scores quite highly on all 4 metrics, we observe a high rate of named entity hallucination in the converted outputs.
For instance, for the input utterance ``Hannah: Hey, do you have Betty's number?'', the no context model outputs ``Hannah asks John if he has Betty's number.'' 
However, the other conversational partner in this dialogue is ``Amanda,'' not ``John.''
Because the gold perspective shifts were annotated with the full conversation available for reference, this model often hallucinates to fill in named entity slots that it does not have the context to resolve.

\begin{table*}[t]
\centering
\begin{tabular}{l|llll}
Approach                     & ROUGE-1 & ROUGE-2 & ROUGE-L & BARTScore \\ \hline
\textsc{PS only}    &    63.57       &    40.74     &     62.04    &    -2.36       \\
\textsc{formality + PS} &    62.00      &  39.14       & 60.38        &   -2.37 \\
\textsc{formality only} &   51.25      &  22.12       &  49.96       &  -2.57         \\
\textsc{rules-based heuristic}        &  61.77       &  35.93       &   55.34      &   -2.80      \\
\textsc{heuristic  + formality}      &   56.98      &  31.91   &  55.72    &   -2.59    \\
\end{tabular}
\caption{Scores for each of the perspective shift models.}
\label{tab:PSscores}
\end{table*}

By contrast, the \emph{conversation-level} model has the clear advantage of referencing the entire conversation at generation time.
However, the model does not have a requirement to produce the same number of lines as the input and must learn this property during training.
We conjecture that this is the reason for this model's relatively weak performance relative to the \emph{left and right context} model. 
Additionally, if the model generates more or less lines than the input dialogue, this can be a conflating factor in the extractive summarization example we discuss in Section \ref{sec:extractive}.
If the model generates less lines than the input, it has performed some part of the summarization process by abstracting the input into a shorter output; if it has generated more lines than the input, it has produced a harder problem for the extractive summarization system by creating more lines to choose the summary from.
Because of this model's weaker performance and this conflating factor, we restrict our remaining experiments in this paper to models that perspective shift one utterance at a time.

The model with \emph{left context only} mimics how a human might read the conversation for the first time, from top to bottom. 
This choice of model also imposes the constraint that the output is the same number of lines as the input, as desired.
However, the dialogues frequently contain cataphora, especially in the start of the conversations, where the first speaker may be addressing a second speaker who has not yet spoken. 
For instance, in the example ``Hannah: Hey, do you have Betty's number?'', this is the first utterance of the dialogue. 
A model with only \emph{left context} cannot resolve the word ``you'' here any better than the \emph{no context} model.

The \emph{left and right context} model addresses this concern by providing the full conversation as input, but restricting the output generation to a perspective shift for a single (marked) utterance. 
This imposes the output length constraint without sacrificing contextual information.
This model performs best on all 4 metrics. 
As the scores for \emph{left and right context} and \emph{no context} models are relatively close, we conduct a human evaluation comparing these two cases. In our blind comparison of 22 conversations, the \emph{left and right context} model was preferred over the \emph{no context} model 86\% of the time (2 annotators, Cohen's kappa 0.62). 

The \emph{conversation-level} model may be a good choice for some applications, where output length is less important to the downstream task. 
This model has a higher degree of abstractiveness, which can lead to increased fluency but also increased hallucination.
For tasks where this is a concern, the \emph{left and right context} model achieves reasonable fluency while adhering more closely to the task, as measured by the automatic metrics.

\subsection{Formality and Perspective Shift}

\paragraph{Approaches}

We observe that the perspective shifting task requires a high degree of formalization. We consider several models ranging from simple rule-based approaches to those relying on an external formalization dataset in order to better understand the role of formalization in perspective shifting. The external dataset we consider is the Grammarly Yahoo Answers Formality Corpus (GYAFC) \cite{rao-tetreault-2018-dear}: a dataset of approximately 100,000 lines from Yahoo Answers and formal rephrasings of each line.

Our core method is the BART model trained under the \textit{left and right context} formulation (\textsc{PS only}).

We also consider a heuristic baseline (\textsc{rules-based heuristic}). 
For each message, we prepend the speaker's name and the word ``says'' to the utterance. 
We replace each instance of the pronoun ``I'' in the message with the speaker's name. 
After observing that most messages are not well-punctuated, we also append a period to the end of each utterance.
While this heuristic is simple and ignores many pronoun resolution conflicts, it has the clear advantage of being highly efficient.

We incorporate the GYAFC corpus as part of our training regime by finetuning on the formalization task prior to finetuning on perspective shift (\textsc{formality + PS}). 

Finally, we perform an ablation by finetuning BART for formalization on the GYAFC corpus, then attempting zero-shot transfer to the perspective shifting task.
As input for this model at test time, we provide either the original dialogues (\textsc{formality only}) or the output of the rules-based heuristic (\textsc{heuristic+formality}).

\paragraph{Results}

\begin{table*}[t]
\centering
\begin{tabular}{p{0.3\linewidth} | p{0.7\linewidth}}
Approach                     & Perspective shifted output \\ \hline
\textsc{PS only}        &     Igor tells John that he has so much to do at work and he is so demotivated.                       \\
\textsc{formality + PS} &             Igor says shit, he has so much to do at work and he is so demotivated.               \\
\textsc{formality only} &          I've got so much to do at work and I'm so demotivated.                  \\
\textsc{rules-based heuristic}        &  Igor says Shit, Igor has got so much to do at work and Igor is so demotivated.       \\
\textsc{heuristic  + formality} & Igor says, "Shit, Igor has so much to do at work and Igor is so demotivated." \\
\textsc{gold} & Igor has too much work and too little motivation.
\end{tabular}
\caption{Sample outputs for each of the perspective shift models for the input utterance ``Igor: Shit, I've got so much to do at work and I'm so demotivated.``}
\label{tab:PSoutputs}
\end{table*}

We evaluate each approach on ROUGE 1/2/L \cite{lin-2004-rouge} and BARTScore \cite{BARTscore}.
The scores are in Table \ref{tab:PSscores}, and example outputs are in Table \ref{tab:PSoutputs}. 

At first glance, perspective shift is a task closely related to formalization.
However, the addition of formalization data leads to a slight decrease in model performance. 
This may be due to the formality data biasing the model toward minimal rephrasings, as there is generally relatively low edit distance between the informal and formal sentences in the formality corpus used \cite{rao-tetreault-2018-dear}.
However, for high performance on perspective shift, the addition of clarifying words, emotion attributions, and pronoun substitutions is necessary; these are high-edit-distance operations that are not observed frequently in the formality data.

Formalization without any additional training for perspective shift is, as expected, far weaker than the perspective-shift-only model. 

The rules-based heuristic appears competitive in ROUGE, but both the BARTScore scores and a manual inspection of the output reveal that this approach is lacking.

In the next section, we explore a downstream task: extractive summarization. For all extractive summarization experiments, we use model-generated perspective shift data from the perspective-shift-only model. We train a model on only the validation-set PS data to generate perspective shifts for the train set of SAMSum, and we train a model on only the train-set PS data to generate perspective shifts for the validation set of SAMSum. 

\section{Application: Extractive summarization}
\label{sec:extractive}
In the extractive summarization setting, phrases or sentences are taken directly from the input and composed into a summary.
This is a clear failure case for dialogue, where sentences in the input are in first person and often pose questions or corrections to previous utterances; knowledge of other speakers in the dialogue can be necessary to contextualize the information.
Summaries should present an overview of a conversation that incorporates global contextual information; generally, these summaries are also expected to be in third person. 

Extraction over a perspective-shifted dialogue does not suffer from many of the same problems as extraction over an original dialogue.
The text in a perspective shifted dialogue is in formal third person, which matches the desired style of the summary text.
While individual sentences of the perspective shifted dialogue correspond directly to individual utterances in the dialogue, the coreference resolution involved in the perspective shift step means that these sentences are less interdependent than the dialogue turns. 
In many respects, perspective shift should make the task of dialogue summarization easier.

\subsection{Oracle Extraction after Perspective Shift}

\paragraph{Methods}

This intuitive result is confirmed by the performance of an oracle extractive model.
Given both the input and the summary, the oracle model is tasked with choosing a combination of $k$ utterances from the input to maximize ROUGE. 
\begin{table*}[t]
\centering
\begin{tabular}{l|lll}
Method                      & ROUGE 1        & ROUGE 2        & ROUGE L        \\ \hline
Extractive: Longest-3 utterances        & 32.46          & 10.27          & 29.92          \\
Extractive: Oracle over original SAMSum & 45.89          & 16.35          & 34.80          \\
Extractive: Oracle over PS SAMSum       & 50.63          & 21.40          & 39.11          \\
Abstractive: BART-large finetuned        & 52.863 $\pm$ 0.531 & 28.577 $\pm$  0.470 & 43.727 $\pm$  0.772
\end{tabular}
\caption{Performance of oracle extractive models as compared to the best extractive baseline from the SAMSum paper (longest-3) and a competitive abstractive system (BART-large, averaged over 5 random restarts).}
\label{tab:oracles}
\end{table*}
Table \ref{tab:oracles} shows the performance of an oracle extractive model over the original SAMSum dialogues and the perspective shifted versions. 
For comparison, a simple extractive baseline---choosing the three longest utterances---and a strong abstractive model are also reported. 

\begin{table*}[t]
\centering
\begin{tabular}{lllll}
Training data & Test data & ROUGE 1 & ROUGE 2 & ROUGE L \\ \hline
CNN/DM        & SAMSum    & 35.00     & 12.09     & 30.76        \\
CNN/DM        & PS SAMSum &   37.12      &  13.14       &    31.49     \\ 
SAMSum        & SAMSum    &   32.19    & 9.86       &   28.52       \\ 
PS SAMSum     & PS SAMSum & \textbf{39.58}    &   \textbf{15.03 }    & \textbf{  33.94   }    
\end{tabular}
\caption{Results for zero-shot (rows 1--2) and supervised extractive models (rows 3--4).}
\label{tab:SUMscores}
\end{table*}

\paragraph{Results}
Clearly, the potential (best-case) performance of a model over the perspective shifted dialogues is better; the oracle scores over perspective shifted dialogues even approach the scores of the abstractive model.

\subsection{Zero Shot and Supervised Extractive Summarization}

\paragraph{Train/Test Regimes}

A common summarization domain is news articles due to the relatively wide availability of data. 
We use an extractive summarizer trained on the CNN/DM news summarization corpus \cite{nallapati-etal-2016-abstractive}: the model PreSumm, introduced by \citet{liu-lapata-2019-text}.
PreSumm uses BERT \cite{devlin-etal-2019-bert} as a backbone to learn a representation of each sentence; additional document-level transformer layers predict whether each sentence should be included in the extractive summary. 
We apply this model for zeroshot summarization\footnote{Here we use the term \textit{zeroshot} to refer to having zero examples in the \textit{target} domain of dialogue summarization.} over the original SAMSum dialogues and over a perspective-shifted version of the dialogues.
We also consider the fully supervised case; we train models using the PreSumm architecture for extraction over the original SAMSum dialogues and over the perspective-shifted dialogues. 

\paragraph{Results}

Results across all models are in Table \ref{tab:SUMscores}.
The zeroshot model scores higher than the supervised model for SAMSum, which at first appears unintuitive.
We credit this to 2 factors.
First, the training dataset for CNN/DM is approximately 21x more training examples than SAMSum train set, allowing the model increased generalizability to an unseen test set.
Second, the summaries in the CNN/DM dataset are often several sentences, while the summaries in the SAMSum dataset tend to be a single sentence. 
The CNN/DM model's bias toward longer summary length may artificially inflate ROUGE scores, as the model selects more utterances for the output.
Despite these factors, the supervised model over perspective shifted data outperforms the zeroshot model over the same data.

Perspective shift is useful as an operation to bring the dialogue domain closer to the news domain.
This drastically improves zero-shot transfer. 
The zero-shot model over perspective shifted data performs better than the fully supervised model trained over the original dialogues.
In a low-data setting, where annotating the entire dataset for summarization may be cost-prohibitive, perspective shift can serve as an alternative annotation goal.
The perspective shift model used to generate the test data in Table \ref{tab:SUMscores} was trained on 545 dialouges (with a validation set of 137 dialogues); by contrast, annotating the entire train and validation sets for summarization would require annotating 15,550 conversations, a more than 20fold increase in annotation effort.

\subsection{Analysis of Hallucination}
\label{sec:benefits-of-ps}

One of the oft-cited benefits of extractive summarization is that models that copy text directly from the input are less likely to present factually incorrect summaries \cite{ladhak-etal-2022-faithful}.
Clearly, perspective shifting introduces a rephrasing step into the summarization pipeline.
A natural concern is the potential presence of ``cascading errors''---where errors in the perspective shifting process lead to hallucinatory extractive summaries. 
We randomly select 100 conversations and associated summaries from the perspective-shift-then-extract model and a standard abstractive finetuned BART model. 
We then ask 2 annotators to label each for faithfulness-- ranking the summary -1 if it describes information that contradicts the conversation, 0 if it contains information that cannot be verified or falsified by the conversation, and 1 if all information stated in the summary is derived from the conversation.
Cohen's kappa between these two annotators was 0.49, with annotators disagreeing on 12.6\% of summaries. 
For cases where the annotator scores differ, we ask a 3rd annotator to label the conversation and choose the majority opinion.
Results of this evaluation are in Table \ref{tab:human-eval}.

\begin{table}[t]
\begin{tabular}{l|ll}
Model              & \% contradict & \% hallucinate \\ \hline
extractive PS     &      3 \%          &           5 \%      \\
abstractive  &           18 \%       &     22 \%            
\end{tabular}
\caption{Human evaluation results for the extractive model over perspective shifted data and the abstractive model over original SAMSum. }
\label{tab:human-eval}
\end{table}

While perspective shift introduces some hallucinations into the dataset, the rate of hallucination is far lower than for abstractive models.

In the 100 randomly selected conversations, we observe 5 hallucinations introduced by the perspective shifting operation that influence the downstream summaries. 
In the same conversational sample, 22 summaries from the abstractive model contain hallucinations, commonly in the form of incorrectly attributing actions to entities or negating the implications from the original conversations. 
Here, we define a \textit{hallucination} as a statement that is not verified by the source text. Some hallucinations are directly contradictory with the source material (\textit{contradictions}); there are 3 such contradictions in the extractive summaries and 18 such contradictions in the abstractive summaries. 

\subsection{Fluency}
Extractions from text message dialogues are not normally conducive to forming a fluent summary.
Each message has its own speaker who may use first person pronouns.
Additionally, messages often contain slang or emojis, which are not appropriate for a formal summary. 
Perspective shifted dialogues are more formally written and describe the conversation from a single frame of reference.

To compare the fluency of extraction from original dialogues and perspective shifted dialogues, we calculate the perplexity of the output summaries for each model.
We measure perplexity using GPT-2 \cite{gpt2}, which is not used to generate any of the outputs. 
The extractions from the perspective shift dialogues have an average perplexity of 31.07, while the extractions from the original dialogues have an average perplexity of 48.77.
Example outputs from each model are in Appendix \ref{sec:appendix-model-outputs}. 

Similarly to extract-then-abstract systems, perspective shifting represents a compromise between the strong faithfulness of extraction and the improved fluency of abstraction. 
\section{Discussion}

Another possible application of perspective shift for summarization is in query-specific summarization, where there is not a single canonical summary at training time.
Instead, a relevant span is selected and summarized based on a user query.
Query-specific summarization has been applied to dialogue-based domains, such as meeting summarization \cite{zhong-etal-2021-qmsum}. In these domains, we conjecture perspective shift may make the choice of an extractive summarization model feasible, allowing for greater interpretability and faithfulness of outputs.

Perspective shift also appears to be a less effortful task for annotators than summarization. We ask a crowdworker to perform perspective shift and summarization annotation for 5 hours each over different sets of dialogues. The annotator gave this unsolicited feedback:
\begin{quote}
{\small [Summarization] is a completely different task in that it takes a lot more mental capacity, paraphrasing complete conversations into a concise synopsis. I need to take a break!}\footnote{Quote used with permission.}
\end{quote}
\noindent This annotator was able to summarize conversations at a faster hourly rate than perspective shifting, but reported that the perspective shift task was more enjoyable. 

We discuss perspective shift for different dialogue subdomains briefly in Appendix \ref{sec:appendix-other-datasets}.
\section{Related Work}

\textbf{Style Transfer}
The most similar style transfer task is formalization, which has attracted attention as a standardization strategy for noisy user-generated text. 
Formalization can be performed as a supervised learning task, and supervised approaches often use the parallel sentence pairs from Grammarly Yahoo Answers Formality Corpus \cite{rao-tetreault-2018-dear}.
More commonly, however, formalization is performed as a semi-supervised (\citet{chawla-yang-2020-semi}, \citet{liu-etal-2022-semi}) or unsupervised objective (\citet{krishna-etal-2020-reformulating}, \citet{unsup-probabilistic}). 

Another related style transfer task is the 3rd to 1st person rephrasing task proposed by \citet{granero-moya-oikonomou-filandras-2021-taking}.
This task is evaluated with exact-match accuracy, and their best model achieves 92.8\% accuracy on the test set. 
We conjecture perspective shift is a more difficult task because of its many-to-one nature, as well as the additional emotion attribution and formalization required. 

\textbf{Speaking-style transformation}
Speaking-style transformation is a task which seeks to transform a literal transcription of spoken speech to one that omits disfluencies, filler words, repetitions, and other characteristics of speech that are undesirable in written text.
This task attracted notice particularly in the statistical machine translation community (\citet{sst-article}, \citet{sst-smt}). 
Other work has focused on the opposite direction: converting written text into a style more favorable for audio listening \cite{abu-jbara-etal-2011-towards}.
More recent work has focused on end-to-end modeling without this intermediate step (\citet{salesky-etal-2019-fluent}, \citet{jamshid-lou-johnson-2020-end}).
This task differs from perspective shift in several respects: the focus of speaking-style transformation is on removing disfluencies, whereas perspective shift aims to preserve information that may be conveyed by the informal style of text; perspective shift requires complex coreference resolution and utterance contextualization, while speaking-style transformation leaves references unresolved; and perspective shift is applied to text post-hoc, while speaking style transcription may be performed over transcripts or in an online setting, during speech transcription.

\textbf{Dialogue summarization}
Much of the prior work on dialogue summarization has focused on incorporating specific features of dialogue into the model architecture or generation process:
\newcite{narayan-etal-2021-planning} incorporate an ordered list of entities to mention in the summary; \newcite{liu-etal-2021-coreference} add coreference resolution information (labeled by a different model) into the input; \newcite{khalifa-etal-2021-bag} add span markers for negations. 
Other work focuses on modeling dialogue structure: \newcite{lei-etal-2021-finer-grain} model the semantic structure of each speaker's contributions individually to build a global semantic structure; \newcite{chen-yang-2021-structure} extract conversational structure from several views to feed into a multi-view decoder.  
Another approach to modeling the differences between dialogues and well-structured text is to use auxilary tasks during training  \cite{liu-etal-2021-topic-aware}.
In work concurrent with this paper, \newcite{fang-etal-2022-spoken} propose a narrower utterance rewriting task for dialogue summarization, swapping some pronouns in the text for speaker names; however, this task does not allow for full rephrasings of the text or produce output that is in third person, making it unsuitable for extractive summarization. 

\textbf{Domain adaptation for summarization}
Another popular direction for dialogue summarization is domain adaptation to dialogue, primarily by pretraining models on additional dialogue data.
\newcite{khalifa-etal-2021-bag} pretrain BART on informal text before training on SAMSum, observing improvement when pretraining on dialogue corpora but not when training on Reddit comments. 
\newcite{yu-etal-2021-adaptsum} study the effectiveness of adding an additional phase of pretraining to improve domain adaptation, in which they either train on a news summarization task, continue pretraining (using the standard reconstruction loss) for an in-domain dataset, or continue pretraining on a smaller dataset of unlabeled input dialogues from the training set.
\newcite{zou-etal-2021-low}  pretrain an encoder on dialogue and a decoder on summary text separately before training the two together on a summarization objective. While these approaches improve performance on dialogue summarization, particularly in a lower-resource setting, they largely require pretraining at a large computational cost.

\section{Conclusion}
Perspective shift is a new, non-trivial style transfer task that requires incorporation of coreference resolution, formalization, and emotion attribution.
This paper presents a preliminary dataset for this task that includes interpretation of the meanings of conventional text abbreviations, emojis, and emoticons. 
The baselines presented in this paper are sufficient for downstream performance on a summarization task, but may be further improved by modeling the unique challenges of this task direction. 

In addition to being a challenging task, perspective shift is a useful operation for dialogue summarization. 
Perspective shift can act as a tool for domain adaptation by shifting dialogue into a form more similar to common summarization domains (e.g. news).
For extractive systems, dialogue summarization is largely infeasible because outputs will not be fluent.
Perspective shift allows for fluent extractive summaries.
This differs from a more traditional extract-then-abstract approach because the ``abstraction'' (perspective shifting) step can benefit from the full document context.
In a domain such as dialogue, where many utterances are strongly conditioned on the prior context of the conversation, this allows for more faithful rephrasings. 
When coupled with an extractive system, this perspective-shifting-based paradigm allows for the creation of more interpretable, less hallucinatory summarizations when compared to an abstractive model.

Other potential applications of perspective shift include direct application in abstractive summarization; in related tasks such as key point analysis \cite{bar-haim-etal-2020-arguments}, which often rely on dialogues as inputs; and for summarization of texts which contain partial dialogues, such as novels. 
The general strategy of transforming the input to adapt to a new domain rather than changing the model or pretraining paradigm is a promising direction because of the ease of annotation and relatively low computational cost. 

\section{Limitations}

Perspective shift requires the modeling of informal language, a challenging task. 
The meaning of informal language can vary across communities \cite{jorgensen-etal-2015-challenges}, age groups \cite{Rickford2013GirlzIW}, and time \cite{lifelong-pretraining}, making generalization of these results more difficult.
This is also an inherently lossy conversion; though we take steps to minimize the loss of emotion and stance information, the nuances of this information may still be discarded. 

The perspective shift process also discards most of the discourse information available in the original dialogue. 
By performing perspective shift prior to dialogue summarization, we take a simplistic view of dialogue as a linear collection of first-person statements without considering underlying structure. 
While this approach proved effective, we believe that the best possible performance on this task may be constrained by this simplifying assumption.

\bibliography{anthology,custom}
\bibliographystyle{acl_natbib}

\newpage
\clearpage

\appendix
\onecolumn

\section{Model-generated perspective shift outputs}
\label{sec:appendix-PS-outputs}
\begin{table*}[h!]
\begin{tabular}{p{0.48\linewidth} | p{0.48\linewidth}}

Original                                    & Perspective shifted                                        \\ \hline
Crystal: \textless{}file\_photo\textgreater{}  & Crystal sends a photo.\\
Irene: He's so big! & Irene says that he is so big.\\
Crystal: \textless{}file\_photo\textgreater{} & Crystal sends a photo.\\
Crystal: I know right! & Crystal agrees with Irene.\\
Irene: and so cute! & Irene adds that it is so cute.\\
Crystal: he got so big he doesn't fit his clothes anymore & Crystal says \textbf{his} son got so big that he does not fit his clothes anymore.\\
Irene: time to go shopping with my little boy <3 & Irene tells her it is time to go shopping with her little boy and sends her love.\\ 
Crystal: yeah im just gonna go brankrupt & Yeah, Crystal is just going \underline{brankrupt}.\\
Irene: Let me take him, I also promise to buy him something & Irene tells Crystal that she will take her son shopping with her, and that she also will buy him something.\\
Crystal: you really wanna do that? & Crystal asks if \underline{Irea} really wants to do that.\\
Irene: why nont? I'm his aunt! & Irene asks why not, she's his aunt.\\
Crystal: well yeah it's just such a drag & \textit{Yeah}, Crystal says, it is a \textit{drag}.\\
Irene: you were always a bore when shopping :P just let me take the little man  & Irene tells Crystal that she was always a bore when shopping, and that she should just let her take the little man. \\
Irene: well have fun! & Irene tells her to have fun.\\
Crystal: ok  &Crystal agrees.\\

\end{tabular}
\caption{A randomly selected example conversation from the SAMSum dataset with model-generated perspective shift. Pronoun errors are \textbf{bolded}; formality errors are \textit{italicized}, and misspellings are \underline{underlined}}
\label{tab:modelPS-ex1}
\end{table*}

\begin{table*}[h!]
\begin{tabular}{p{0.48\linewidth} | p{0.48\linewidth}}

Original                                    & Perspective shifted                                        \\ \hline

Maria: hey guys!  & Maria greets the group enthusiastically. \\ 
Maria: everything ready for the conference? & Maria asks if everything is ready for the conference. \\ 
Kate: yes, almost & Kate says yes, almost. \\ 
Tommy: I think we will have a good panel & Tomy thinks they will have a good panel. \\ 
Sam: I really hope there will be some people & Sam tells them that he really hopes there will be some people. \\ 
Tommy: we should have good audience & Tommy thinks they should have a good audience. \\ 
Maria: Are you preparing a power point? & Maria asks if they are preparing a \underline{power point.} \\ 
Tommy: I'll have a prezi & Tommy says he will have a \underline{pre-conference.} \\ 
Maria: you're paying for it? & Maria asks if Tommy is paying for it. \\ 
Tommy: it's good, worth the money & Tommy thinks it is good and that it is worth the money. \\ 
Maria: how much is it? & Maria asks how much it is. \\ 
Tommy: I believe \$10 a month for academics & Tommy replies that he believes it is \$10 a month for academics. \\ 
Maria: not that bad  &  Maria says it's not that bad. \\

\end{tabular}
\caption{A randomly selected example conversation from the SAMSum dataset with model-generated perspective shift. Pronoun errors are \textbf{bolded}; formality errors are \textit{italicized}, and misspellings are \underline{underlined}}
\label{tab:modelPS-ex2}
\end{table*}

\clearpage
\newpage

\section{Summarization model outputs}
\label{sec:appendix-model-outputs}

\begin{table*}[h!]
\begin{tabular}{ll}

\textbf{Original conversation} \\ \hline
Hannah: Hey, do you have Betty's number? \\
Amanda: Lemme check \\
Hannah: \textless{}file\_gif\textgreater{} \\
Amanda: Sorry, can't find it.\\
Amanda: Ask Larry\\ 
Amanda: He called her last time we were at the park together\\
Hannah: I don't know him well\\
Hannah: \textless{}file\_gif\textgreater{}\\
Amanda: Don't be shy, he's very nice\\
Hannah: If you say so..\\
Hannah: I'd rather you texted him\\
Amanda: Just text him \includegraphics[height=1.7\fontcharht\font`\D]{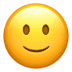} \\ 
Hannah: Urgh.. Alright\\
Hannah: Bye\\
Amanda: Bye bye\\
\hline
\textbf{Original extraction [zeroshot]:} \\
Hannah: Hey, do you have Betty's number? Hannah: I'd rather you texted him Amanda: He called \\her last time we were at the park together
\\ \hline
\textbf{Original extraction [supervised]:} \\
Hannah: Hey, do you have Betty's number? Amanda: He called her last time we were at the park \\together Amanda: Don't be shy, he's very nice \\ \hline
\textbf{Perspective shift extraction [zeroshot]:} \\
Hannah asks Amanda if she has Betty's number. Amanda says she needs to check. Amanda \\ tells Hannah not to be shy, and that he is very nice.\\ \hline
\textbf{Perspective shift extraction [supervised]:} \\
Amanda says that Larry called Betty last time they were at the park together. Hannah asks\\ Amanda if she has Betty's number. Amanda tells Hannah to ask Larry.  \\ \hline
\textbf{Abstractive summary:} \\
Hannah is looking for Betty's number. 
Amanda tells Hannah to ask Larry, who called Betty last \\time they were at the park together. \\ \hline

\textbf{Gold:} \\
Hannah needs Betty's number but Amanda doesn't have it. She needs to contact Larry. \\
\end{tabular}
\caption{An example conversation from the SAMSum dataset with summmaries from each model presented, as well as the gold summary.}
\label{tab:summaries1}
\end{table*}
\begin{table*}[h!]
\begin{tabular}{ll}

\textbf{Original conversation}  \\ \hline
Dima: hello! \\
Nada: hey girl, what's up?\\
Dima: I'm in a huge trouble, my laptop is broken and I have to deliver a translation tomorrow \\@9 \includegraphics[height=1.7\fontcharht\font`\D]{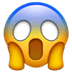}\includegraphics[height=1.7\fontcharht\font`\D]{emojis/face-screaming-in-fear.png}\\
Nada: fuck what happened??\\
Dima: the stupid cat spilled coffee on it \includegraphics[height=1.7\fontcharht\font`\D]{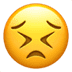}\includegraphics[height=1.7\fontcharht\font`\D]{emojis/perservering-face.png} freaking out \\
Dima: you still have your old laptop? is it possible to lend it to me please? \\
Nada: no sorry, I've given it to my brother - but you're lucky! I've taken these two days off so \\you can take mine \\
Dima: ooh man! thank you sooo much!!! if it weren't for Trados, I wouldn't be panicking :(\\ 
Nada: no worries, it happened... but I always think about this... like man, we need some\\ back up laptops!\\
Dima: I know! but I always change my mind and spend the money elsewhere lol\\
Nada: yeah, but it's like our only tool! so we need to invest in it\\
Dima: yup, true !\\
Dima: can I come in an hour to pick it up?\\
Nada: yes :) ttyl!\\
\hline
\textbf{Original extraction [zeroshot]:} \\
Nada: hey girl, what's up? Nada: fuck what happened?? Dima: I'm in a huge trouble, my laptop \\is broken and I have to deliver a translation tomorrow @9 \includegraphics[height=1.7\fontcharht\font`\D]{emojis/face-screaming-in-fear.png}\includegraphics[height=1.7\fontcharht\font`\D]{emojis/face-screaming-in-fear.png}
\\ \hline
\textbf{Original extraction [supervised]:} \\
Dima: I'm in a huge trouble, my laptop is broken and I have to deliver a translation tomorrow \\@9 \includegraphics[height=1.7\fontcharht\font`\D]{emojis/face-screaming-in-fear.png}\includegraphics[height=1.7\fontcharht\font`\D]{emojis/face-screaming-in-fear.png} Dina: the stupid cat spilled coffee on it \includegraphics[height=1.7\fontcharht\font`\D]{emojis/perservering-face.png}\includegraphics[height=1.7\fontcharht\font`\D]{emojis/perservering-face.png} I'm freaking out! Dima: you still have your\\ old laptop? is it possible to lend it to me please?\\ \hline
\textbf{Perspective shift extraction [zeroshot]:} \\
Dima says hello to Nada. Nada asks Dima what's up. Dima is in huge trouble, her laptop is broken\\ and she has to deliver a translation tomorrow.\\ \hline
\textbf{Perspective shift extraction [supervised]:} \\
Dima is in huge trouble, her laptop is broken and she has to deliver a translation tomorrow. Dima \\asks if he can pick up the laptop in an hour. Nada has taken two days off so Dima can take her \\laptop.. \\ \hline
\textbf{Abstractive summary:} \\
Dima's laptop is broken. He has to deliver a translation tomorrow at 9. Nada lent her laptop \\to her brother. She took two days off. Dima will come in an hour to pick up Nada's laptop. \\ \hline
\textbf{Gold:} \\
Dima's laptop is broken, as her cat spilled coffee on the laptop. Dima is worried, because she has \\to deliver a translation for Trados tomorrow. Dima will come to Nada in an hour to borrow Nada's \\laptop. \\
\end{tabular}
\caption{An example conversation from the SAMSum dataset with summmaries from each model presented, as well as the gold summary.}
\label{tab:summaries2}
\end{table*}

\clearpage
\newpage
\twocolumn

\section{Applicability of Perspective Shift for other datasets}
\label{sec:appendix-other-datasets}

While we present the perspective shift task using text message conversations as an example, there are a wide variety of subdomains within dialogue. We apply the perspective shift operation to two other domains-- roleplaying game transcripts and media interviews-- using the model trained only on data from the text message conversation domain.

While the model effectively perspective shifts most short utterances, the largest issues we observed in inspection of these outputs are as follows:
\begin{enumerate}
    \item \textbf{Long utterances:} The perspective shift model performs poorly when utterances are very lengthy, as this is very uncommon in the SAMSum dataset (average utterance length: 8.4 words). This leads to repetition and denigrated performance, especially when several long utterances occur in sequence.
    \item \textbf{Domain differences in formatting:} Differences such as multi-word speaker names or adding sound effects in parentheses are not captured effectively by the model, as they were not encountered at training time. 
\end{enumerate}

While we leave improving perspective shift over long outputs to future work, we provide examples of perspective shifts from two different domains, to demonstrate these potential pitfalls for other researchers. 
These are model-generated perspective shifts, generated using the model trained only on perspective shift for SAMSum dialogues.
\subsection{CRD3}
CRD3 is a dataset of Dungeons \& Dragons roleplaying game transcripts \cite{rameshkumar-bailey-2020-storytelling}. Dungeons \& Dragons is a collaborate roleplaying game where multiple players describe the actions and dialogue of their characters as the team explores an open-ended world. While each session of the game consists of several thousand turns of dialogue, the CRD3 dataset sections the sessions into smaller chunks with aligned summaries. For brevity, we present only a chunk of a session, in Table \ref{tab:crd3}. The SAMSum perspective shift model serves as a reasonable baseline for this dataset, though in-domain data would likely further improve performance.

\subsection{MediaSum}
MediaSum is a dataset of NPR and CNN media interview transcripts \cite{zhu-etal-2021-mediasum}.
The average turn length in this dataset is substantially longer-- 37.5 words for the NPR transcripts and 53.1 words for the CNN transcripts.
Correspondingly, the model-generated perspective shift is worse.
The model generates repetitious content in the perspective shift.
The model also performs poorly on multi-word speaker names, which are a rarity in SAMSum as well. 
A snippet of an interview appears in Table \ref{tab:mediaSum}

\clearpage
\onecolumn

\begin{table*}[h!]
\begin{tabular}{p{0.48\linewidth} | p{0.48\linewidth}}

Original                                    & Perspective shifted                                        \\ \hline
MATT: Okay. You take your first step and you watch something drift across the entrance from wall to wall. Some faint glowing figure-- and is gone. & Matt says that when you take your first step and you watch something drift across the entrance from wall to wall, some faint glowing figure-- and then it is gone. \\ 
TRAVIS: (yelling) I don't like it! & Trevor yells that he does not like it. \\ 
MARISHA: Shut up! Oh. & Marisha is shocked. \\ 
TRAVIS: It's in my head! & Trevor says it is in his head. \\ 
MARISHA: Oh, okay. & Marisha agrees. \\ 
TALIESIN: Did it trip my Detect Magic sense at all? & Talyiesin asks if it trip her Detect Magic sense at all. \\ 
MATT: Not magic sense, no. & Matt says it's not a magic sense. \\ 
LAURA: Is it undead, or is it a ghost? & Laura asks if it is an undead or a ghost. \\ 
MARISHA: (whispering) It's Dashilla! & Marisha says it's Dashilla. \\ 
TALIESIN: I'm going to do another detect undead really quickly. & Talesin tells her that he is going to do another one. \\ 
MATT: Okay. Give me the specifications on that one. & Matt agrees and asks for the specifications. \\ 
TALIESIN: All right. Eyes of The Grave. As an action you know the location of any undead within 60 feet of you that isn't behind total cover and isn't protected from divination magic until the end of your next turn. & Talyiesin agrees and mentions that it is the Eyes of the Grave, and that as an action, it can alert the location of any undead within 60 feet or less that isn't covered by total cover and isn't protected from divination magic until the end of the next turn. \\ 
MATT: Okay, got you. How long does it last? & Matt tells her that he got her. \\ 
TALIESIN: Six seconds. & Talesin says it takes six seconds. \\ 
MATT: Okay. We'll say for the purposes of this, this is a reaction to seeing this figure pass by. You definitely get an undead sense from whatever this figure is, and then it merges with the wall and is gone. It appeared to be loosely humanoid in the brief glimpse you saw. & Matt tells her that they will say that for the purposes of this, it is a reaction to seeing the figure pass by, and that you definitely get an undead sense from whatever the figure is, and then it merges with the wall and is gone. \\
TALIESIN: We have undead. & Talyiesin says they have undead. \\ 
TRAVIS: No. & Trevor says no. \\ 
SAM: It's a ghost ship! & Sam says it is a ghost ship. \\ 
MARISHA: Is it going to be your crew? & Marisha asks if it is going to be Ducey's crew. \\ 
TRAVIS: I don't know! & Trevor doesn't know. \\ 
MARISHA: Wait, we're not on your ship anymore. & Marisha asks them to wait.  She tells them that they are not on Ducey's ship anymore. \\ 
TRAVIS: Now we're in the mouth of a beast! Ducey, what's going on, brother? & Trevor tells her that they are now in the mouth of a beast, and asks Ducey what is happening. \\ 
TALIESIN: I don't know, I'm literally blind. I want to light this-- & Talesin doesn't know, she is literally blind. \\ 

\end{tabular}
\caption{A snippet of a conversation from CRD3 with a model-generated perspective shift.}
\label{tab:crd3}
\end{table*}

\begin{table*}[h!]
\begin{tabular}{p{0.48\linewidth} | p{0.48\linewidth}}

Original                                    & Perspective shifted                                        \\ \hline
MADELELEINE BRAND, host: OK, here's some good news on the jobs front for both men and women. A new survey out today from the employment firm Manpower finds that about a quarter of employers will add jobs this summer. That's for adults, but for teenagers this summer's job market is shaping up to be the weakest in more than 50 years. & MADELEINE says that there is some good news on the job market for both men and women. A new survey out today from the employment firm Manpower finds that about a quarter of employers will add jobs to the workforce this summer, but for teenagers, the summer job market is the weakest in more than 50 years.\\
ALEX COHEN, host: So, how do you get your teenage kids not to spend the entire summer glued to the couch? You're about to get some tips from Michelle Singletary. She's Day to Day's personal finance contributor. Hi, Michelle! & Alexo Cohen, host, tells Michelle that the types of jobs that teens tend to go for in retail, fast food, and the like, that they still need people.  He says that oftentimes, listen, you may not get a job at the beginning of the summer, but hold on because in late summer, when some college students are going back to school and perhaps some of the people who lost their\\
MICHELLE SINGLETARY: Hi! & MICHelle says hello.\\
ALEX COHEN, host: So why is the summer job market so hard for teens this year? &  Alex Cooper says that the types of jobs that teens tend to go for in retail, fast food, and the like, they still need people.  He tells Michelle that they've got a tough economy and that they have a lot of college graduates going into the job market and taking jobs that would traditionally go to teens, like in restaurants and retailers, but that for teenagers, the summer job market is the weakest in more than \\
MICHELLE SINGLETARY: Lot of things going on right now. We've got a tough economy. We've got a lot of college graduates going into the market. We have people who are losing their jobs and taking jobs that would traditionally go to teens, like in restaurants and retailers. And we have a lot of older people holding on to their jobs and not retiring because they can't afford to retire. And that puts teens at the end of the line when it comes to these types of jobs. & MICHelle says that there are a lot of things going on in the job market right now, and that they have a tough economy.  She says that they've got a number of college graduates going into the workforce and taking jobs that would traditionally go to teens, like in restaurants and the like, but that for teenagers, the summer job market is the weakest in more than 50 years.\\

\end{tabular}
\caption{A snippet of a conversation from MediaSum with a model-generated perspective shift. The model performs poorly on inputs with several long utterances.}
\label{tab:mediaSum}
\end{table*}

\newpage

\clearpage

\twocolumn

\section{Annotation Information}
\label{sec:appendix-annotation}

Annotators were selected through two processes.
Annotators on Amazon Mechanical Turk were selected after successfully completing a pilot annotation task.
Annotators on UpWork were selected after bidding on the contract and received feedback after their first 5 annotated conversations. 
Mechanical Turk annotators were compensated \$0.33 for annotating one conversation in the pilot round and \$0.90 per conversation annotated in the annotation round, for estimated hourly wages of approximately \$15 per hour\footnote{To arrive at this estimate, we timed one author annotating for 20 minutes and projected this into an hourly annotation rate.}. 
UpWork annotators were paid between \$10 and \$15 per hour, according to the rates they bid for the job. 
A total of 20 Mechanical Turk workers and 6 UpWork workers contributed annotations for this dataset.
All annotations were reviewed and lightly edited by the authors for quality. 
We estimate the cost of recreation for this dataset to be approximately \$900. 

The annotation instructions below were provided to all annotators, along with an accompanying example for each of the 8 instructions and several full conversations annotated as examples.

\subsection{Annotation instruction}

The goal of the annotation is to transform from informal 1st person to formal 3rd person,\textit{ while preserving tone as much as possible. }

Do not include emoticons, slang, or informal language; rewrite the utterance so that someone who grew up pre-AOL-instant-messenger would easily interpret the meaning. Each rewritten utterance should have the speaker’s name in it, but it is not necessary to indicate who is speaking in each sentence. To preserve tone, you can add adverbs or reword sentences slightly. Otherwise, try to keep sentences as close to the original as possible.

Difficult cases should be marked by checking the checkbox in the “Difficult?” column. If the annotation requires substantial thought or you are not confident in your annotation, mark this box and it will be re-reviewed. 
\begin{enumerate}
    \item Each rewritten utterance by a speaker should have that speaker's name in it. Not every sentence in the utterance has to contain the speaker's name if it’s clear from context.
    \item Where possible, prefer inserting speaker names to adding a dialogue tag like ``says'' or ``asks.'' 
    \item Standardize grammar and punctuation. Use adverbs or rephrasing to replace emoticons and text slang.
    \item For attachments sent in the conversation, use the form ``<SPEAKER NAME> sends a <ATTACHMENT TYPE>,'' e.g. ``John sends a file.''
    \item References to ``you'' or ``we'' should be resolved as best you can with the context. If there’s no way to tell who ``we'' refers to, resolve it to ``they.''
    \item Resolve ambiguous references with the context from the rest of the conversation. 
    \item Questions should be rewritten in descriptive form.
    \item When possible, use only speaker names. If the sentence would be disfluent with the speaker name, use the pronouns listed in the conversation header (the line above the conversation) \footnote{These pronouns were generated by applying the Python package \texttt{gender\_guesser} to the speaker names and injecting some degree of randomness; options for speaker pronouns were she/her/hers, he/him/his, or they/them/theirs. We used this approach to avoid asking annotators to make judgement calls about the genderedness of a name. We used an automatic process because the conversations and speakers in SAMSum are synthetic; manual annotation for pronoun use is recommended if using real-world conversations, to avoid misgendering participants during annotation.}. 

\end{enumerate}

\newpage

\end{document}

%% file: mathabbrev.tex
\newcommand{\uv}{\mathbf{u}}

\newcommand{\yv}{\mathbf{y}}